\documentclass[11pt,a4paper]{article}
\usepackage[hyperref]{eacl2021}
\usepackage{times}
\usepackage{latexsym}

\usepackage{microtype}
\usepackage{graphicx}
\usepackage{todonotes}
\usepackage{mathtools, nccmath}
\usepackage{url}
\usepackage{amsfonts}
\usepackage{array}
\usepackage{subcaption}

\usepackage{microtype}

\aclfinalcopy 
\def\aclpaperid{971} 

\title{Scientific Discourse Tagging for Evidence Extraction}

\author{Xiangci Li\textsuperscript{\rm 1}\thanks{~~Work performed while the author is interning at the Information Sciences Institute, University of Southern California} ~~ Gully Burns\textsuperscript{\rm 2} ~~ Nanyun Peng\textsuperscript{\rm 3}\\
  {\tt lixiangci8@gmail.com,gully.burns@chanzuckerburg.com} \\
  {\tt violetpeng@cs.ucla.edu} \\
  \textsuperscript{\rm 1} University of Texas at Dallas, 
  \textsuperscript{\rm 2} Chan Zuckerburg Initiative \\
  \textsuperscript{\rm 3} University of California Los Angeles}

\date{}

\begin{document}
\maketitle
\begin{abstract}
Evidence plays a crucial role in any biomedical research narrative, providing justification for some claims and refutation for others. We seek to build models of scientific argument using information extraction methods from full-text papers. We present the capability of automatically extracting text fragments from primary research papers that describe the evidence presented in that paper's figures, which arguably provides the raw material of any scientific argument made within the paper.
We apply richly contextualized deep representation learning pre-trained on biomedical domain corpus to the analysis of scientific discourse structures and the extraction of ``evidence fragments'' (i.e., the text in the results section describing data presented in a specified subfigure) from a set of biomedical experimental research articles. We first demonstrate our state-of-the-art scientific discourse tagger on two scientific discourse tagging datasets and its transferability to new datasets. We then show the benefit of leveraging scientific discourse tags for downstream tasks such as claim-extraction and evidence fragment detection.
Our work demonstrates the potential of using evidence fragments derived from figure spans for improving the quality of scientific claims by cataloging, indexing and reusing evidence fragments as independent documents.
\end{abstract}

\section{Introduction}

\begin{figure}[t]
  \begin{minipage}[]{0.45\textwidth}
    \begin{tabular}{cc}
    \includegraphics[width=\textwidth]{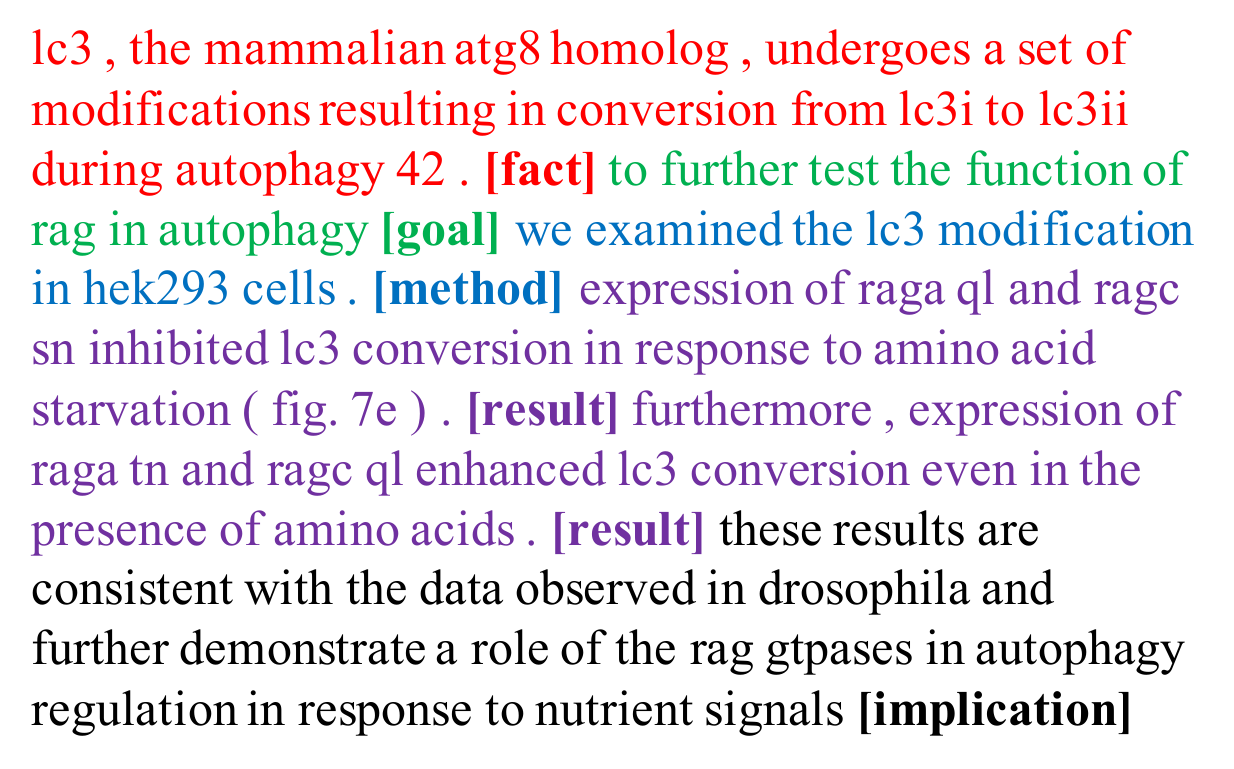}
    \\
  \end{tabular} 
  \end{minipage}
  \\[-0.1cm]
  \vspace{-1.5em}
  \caption{An example paragraph tagged with scientific discourse tags on each clause in SciDT dataset \cite{dasigi2017experiment}. The text is tokenized and converted to  lower case.}
  \label{fig:example}
  \vspace{-0.5em}
\end{figure}

Primary experimental articles (i.e., papers that describe original experimental work) provide the crucial raw material for all other subsequent scientific research. However, the drastically growing number of scientific literature makes it increasingly difficult for domain experts to efficiently utilize them. 
Automatic information extraction from biomedical literature is a crucial step to help researchers to achieve this goal.

Extracting important information from biomedical literature to facilitate and accelerate scientific discovery has been a goal for computational linguistics for some time \cite{hobbs_information_2002}, with the focus of identifying relevant entities, relations, and events from text to populate a knowledge base. 
However, these methods do not take into account the fact that scientific work involves attempting to provide explanations for evidence derived from experiments and is therefore driven principally by authors attempting to convince expert readers that their claims are the ``correct'' explanations for the experimental evidence. 
Thus, an important aspect of building machines capable of understanding scientific literature is first recognizing different \textit{rhetorical components} of scientific discourse, with which we will then be able to distinguish the \textit{observations} made in experiments from their \textit{implications} and distinguish between \textit{claims} supported by evidence and \textit{hypotheses} put forward to prompt further research. It is this goal, of being able to distinguish between the different rhetorical components of scientific discourses so that we can build AI systems to facilitate more accurate analysis and understanding of scientific literature, that motivates our work. 

Scientific discourse tagging is a task that tags clauses or sentences in a scientific article with different rhetorical components of scientific discourses. Figure~\ref{fig:example} shows an example of a paragraph with discourse tags. 
In this work, we leverage a state-of-the-art contextualized word embedding and a novel word-to-sentence attention mechanism to develop a model for scientific discourse tagging that achieves the state-of-the-art performances on two benchmark datasets SciDT~\cite{dasigi2017experiment} and PubMed-20k-RCT~\cite{HSLN} by 6.9\% and 2.3\% absolute F1 respectively. \footnote{\url{https://github.com/jacklxc/ScientificDiscourseTagging}}  
More importantly, we show the strong transferability of our scientific discourse tagger to new datasets by beating the baseline \cite{huang2020coda} via zero-shot prediction on CODA-19 dataset \cite{huang2020coda}. 
Furthermore, we demonstrate the effectiveness of scientific discourse tagging on two downstream scientific literature understanding tasks: claim-extraction and evidence fragment detection, and demonstrate the benefit of leveraging scientific discourse tags information. In particular, we outperform the state-of-the-art claim extraction model \cite{claim-extraction} by 3.8\% F1, and outperform figure span detection baseline \cite{burns2017extracting} by 5\% F1. 

\section{Background and Related Works}

\begin{table}[t]
\begin{center}
    \begin{tabular}{ | c | c |}
    \hline
    \textbf{Type} & \textbf{Definition} \\ \hline
    Goal & Research goal  \\ \hline
    Fact & A known fact, a statement taken \\ & to be true by the author \\ \hline
    Result & The outcome of an experiment  \\ \hline
    Hypothesis & A claim proposed by the author \\ \hline
    Method & Experimental method  \\ \hline
    Problem & An unresolved or \\ & contradictory issue  \\ \hline
    Implication & An interpretation of the results  \\ \hline
    None & Anything else  \\ \hline
    \end{tabular}
    \vspace{1em}
    \caption{Eight label taxonomy defined by \newcite{de2012epistemic}.} \label{tab:taxonomy}
    \vspace{-1.5em}
\end{center}
\end{table}

\noindent \textbf{Problem Formulation.}
We define scientific discourse tagging as a task that labels sentences in a scientific article based on its rhetorical elements of scientific discourse. Formally, a paragraph can be represented as an ordered collection of sequences $\textbf{S}=[S_1, S_2,...,S_n]$, and each element $S_i$ is annotated with a discourse label $L_i\in\{L_1, L_2,...,L_k\}$. Note that $S_i$ may be defined differently in different datasets -- e.g., sentences in the PubMed-RCT dataset~\cite{PubMed-RCT}, clauses in the SciDT dataset composed by \newcite{burns2016automated} and \newcite{ dasigi2017experiment}, and sentence fragments in CODA-19 dataset \cite{huang2020coda}. For conciseness, we refer all these variations as \emph{sentences}. The labels also can be slightly different. For example, in PubMed-RCT, $\textbf{L}=$ \{\emph{objective}, \emph{background}, \emph{methods}, \emph{results}, \emph{conclusions}\}, in CODA-19 \cite{huang2020coda}, $\textbf{L}=$ \{\emph{background}, \emph{purpose}, \emph{method}, \emph{finding/contribution}, \emph{other}\} while in SciDT dataset~\cite{burns2016automated,dasigi2017experiment}, the labels $\textbf{L}=$ \{\emph{goal}, \emph{fact}, \emph{hypothesis}, \emph{problem}, \emph{method}, \emph{result}, \emph{implication}, \emph{none}\} as defined by \newcite{de2012epistemic}. Table~\ref{tab:taxonomy} gives more details about the definitions of the tags.

\subsection{Prior Works on Scientific Discourse Tagging}

\noindent\textbf{Feature-based Scientific Discourse Tagging.}
There has been a significant amount of work aimed at understanding types of scientific discourse. \newcite{teufel1999discourse} and \newcite{teufel2002summarizing} described argumentative zoning, which groups sentences into a few rhetorical zones highlighted by important clauses such as ``in this paper we develop a method for''. \newcite{hirohata2008identifying} used conditional random field (CRF) \cite{lafferty2001conditional} with handcrafted features to classify sentences in abstracts into 4 categories: \emph{objective, methods, results}, and \emph{conclusions}. \newcite{liakata2010zones} defined ``zone of conceptualization'' which classifies sentences into 11 categories in scientific papers and \newcite{liakata2012automatic} used CRF and LibSVM to identify these ``zone of conceptualization". \newcite{guo2010identifying} used Naive Bayes and Support Vector Machine (SVM) \cite{cortes1995support} to compare three schema: section names, argumentative zones and conceptual structure of documents. \newcite{burns2016automated} studied the problem of scientific discourse tagging, which identifies the discourse type of each clause in a biomedical experiment paragraph and composed a dataset for it. They adopted the discourse type taxonomy for biomedical papers proposed by \newcite{de2012epistemic}. The taxonomy contains eight types including \emph{goal, fact, result, hypothesis, method, problem, implication} and \emph{none} as Table~\ref{tab:taxonomy} shows. Most recently, \newcite{cox2017optimized} used the same schema \cite{de2012epistemic} by exploring a variety of methods for balancing classes before applying classification algorithms. 


\noindent\textbf{Deep Learning for Scientific Discourse Tagging.}
Due to the prevalence of deep learning, neural sequence labeling approach using bidirectional LSTM \cite{hochreiter1997long} and CRF (BiLSTM-CRF) \cite{huang2015bidirectional} has been prevailing for classic word-level sequence tagging problems such as named entity recognition (NER), part of speech tagging (POS), and word segmentation~\cite{huang2015bidirectional,peng2015named,peng2016improving,ma2016end,chiu2016named,peng2017multi,wang2017multi,huang2019learning}. Since scientific discourse tagging, which is a sentence-level sequence tagging problem, has one additional dimension of input comparing to word-level sequence tagging problems, an encoder is required to encode word-level representations to clause/sentence-level representations. While one simple way is to pre-compute sentence embeddings from word embeddings \cite{arora2016simple}, there are more sophisticated methods to compute sentence-level embeddings on-the-fly using BiLSTM \cite{HSLN,srivastava2019hierarchical} or attention \cite{dasigi2017experiment}, before feeding them into a clause/sentence-level sequence tagger. Alternatively, as BERT \cite{devlin2018bert} prevails among various natural language processing (NLP) tasks, a simple baseline method is to directly use a BERT-like model's (e.g. SciBERT \cite{beltagy2019scibert}) prefix token ($[CLS]$) representation of each sentence as the sentence representation for classification task \cite{huang2020coda}. In this work, we combine these methods to present a state-of-the-art scientific discourse tagger.

\subsection{Downstream Applications}
\noindent\textbf{Claim Extraction.} 

\begin{figure}
    \includegraphics[width=0.45\textwidth]{figs/claim_example.pdf}
  \vspace{-1.5em}
  \caption{An example abstract with claim sentences highlighted in claim-extraction dataset \cite{claim-extraction}.}
  \label{fig:claim_example}
  \vspace{-0.5em}
\end{figure}

Claim extraction has been extensively studied in various domains. In addition to scientific articles \cite{stab2014argumentation}, previous work has analyzed social media \cite{dusmanu2017argument}, news \cite{habernal2014argumentation, sardianos2015argument} and Wikipedia \cite{thorne2018fact, freard2010role} for a task called \emph{Argumentation Mining} to extract claims and premises. However, there are less attention and dataset available in the biomedical domain. \newcite{claim-extraction} composed a claim-extraction dataset derived from MEDLINE \footnote{\url{https://www.nlm.nih.gov/bsd/medline.html}} paper abstracts, and proposed a neural network model that significantly outperformed the rule-based method proposed by \newcite{sateli2015semantic}. Figure \ref{fig:claim_example} shows an example abstract with the last two sentences annotated as claims.

In this work, we formulate claim extraction~\cite{claim-extraction}
similarly as scientific discourse tagging: $\textbf{S}$ contains sentences and $L_i \in \{0, 1\}$ indicates whether the corresponding sentence is a claim or not.

\begin{figure}
\includegraphics[width=0.45\textwidth]{figs/evidence_fragment_new2.pdf}
  \vspace{-.4cm}
  \caption{An example paragraph of evidence fragment detection. The explicit mention of subfigure codes are underlined. The red lines indicate the borders of the evidence fragments. For each clause, the discourse type as well as the BIO tags indicating ``blocks'' (see Section \ref{sec:methods_evidence_fragment}) are provided. 
  }
  \label{fig:evidence_fragment}
  \vspace{-.7cm}
\end{figure}

\noindent\textbf{Evidence Fragment detection.}
\newcite{burns2017extracting} coined the concept of ``evidence fragments'' as the text section in narrative surrounding a figure reference that directly describes the experimental figure. They composed an evidence fragment detection dataset, and proposed the evidence fragment detection task that tags each clause with semantically referred subfigure codes. They further proposed a rule-based method of using these subfigure codes as anchors to link evidence fragments to European Bioinformatics Institute's INTACT \cite{orchard2013mintact} data records. As a result, INTACT's preexisting, manually-curated structured interaction data can serve as a gold standard for machine reading experiments. 

\newcite{burns2017extracting} formulated the problem into a clause-level tagging problem. Formally, each clause $S_i$ in a paragraph $\textbf{S}=[S_1, S_2,...,S_n]$ is annotated with a set of subfigure codes $f^i=\{f_1^i, f_2^i,...,f_m^i\}$ that each clause is semantically referring to, where the length $m$ can be any non-negative integer. 
Figure \ref{fig:evidence_fragment} shows an illustration of a paragraph of evidence fragment detection annotation. Each clause in the paragraph is associated with a set of semantically relevant subfigures.

\section{Approaches}

\subsection{Scientific Discourse Tagger}
\label{sec:methods_sciDT}
\begin{figure}[t]
  \centering
 \includegraphics[width=.5\textwidth]{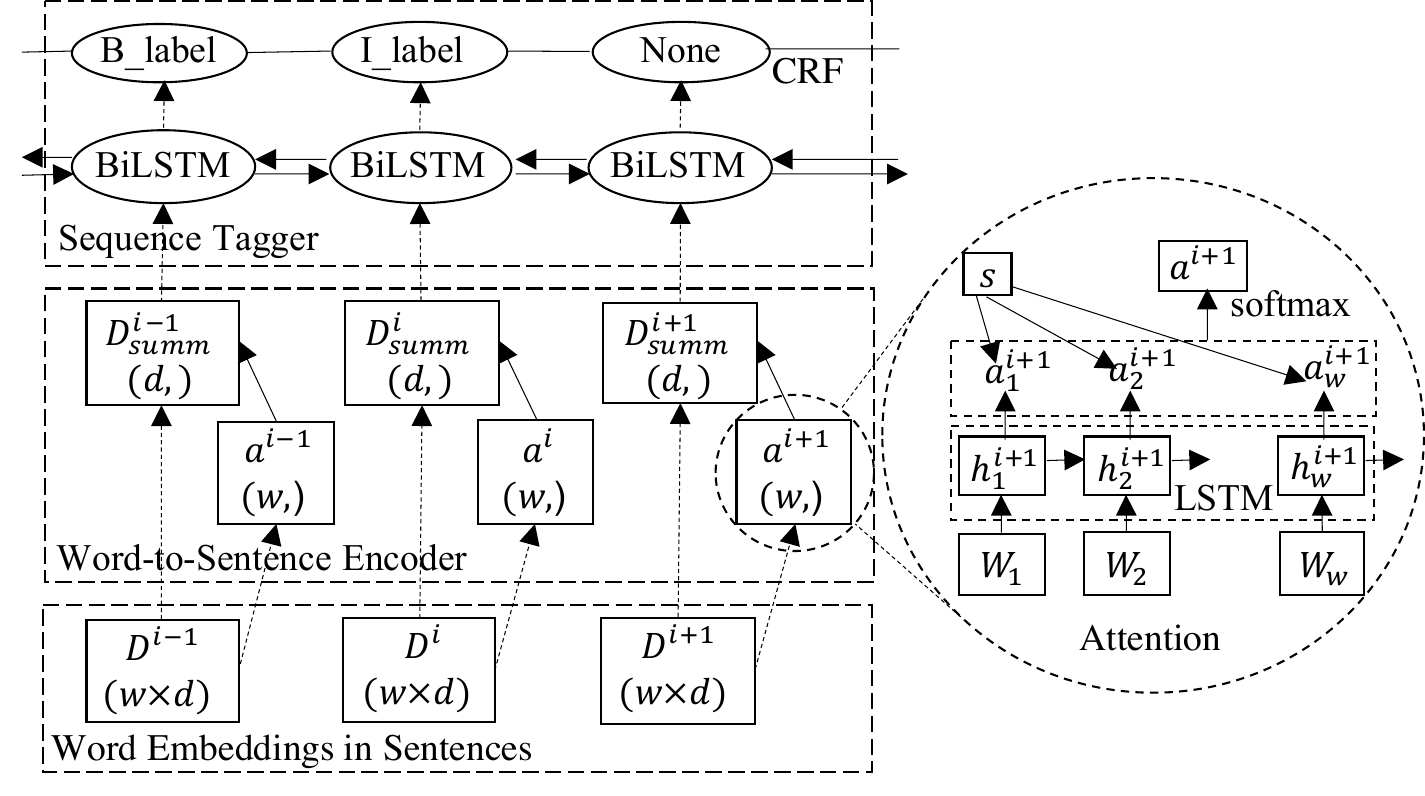}
    \caption{High-level overview of our scientific discourse tagger. Dashed arrows indicate we may apply dropout in those connections.}
    \label{fig:overview}
\vspace{-0.3cm}
\end{figure}

\noindent\textbf{Model Overview.}
We formulate scientific tagging as a sentence level sequence tagging problem. We develop a deep structured model extending~\citet{dasigi2017experiment}, which consists of a contextualized word embedding layer, an attention layer that summarizes word embeddings into sentence embeddings, and a BiLSTM-CRF sequence tagger \cite{huang2015bidirectional} on top of the sentence embeddings for discourse type tagging. Figure~\ref{fig:overview} gives an overview of the architecture. We detail each component in this section.

\noindent\textbf{Embeddings.} 
We explore pre-trained BioGloVe embedding \cite{burns2019building}, BioBERT \cite{10.1093/bioinformatics/btz682} and SciBERT \cite{beltagy2019scibert} embedding, which are GloVe and BERT embeddings trained on the text in biomedical domain.

\noindent\textbf{Sentence Representations via Attention.} 
We observe that only keywords are essential to determine the discourse types, and attention is an appropriate mechanism for emphasizing certain inputs and ignoring others. \newcite{dasigi2017experiment} also explored using an attention mechanism to summarize word representations to sentence representations, however, we propose a new variation of attention mechanism using an LSTM. 
Specifically, we first encode the sentence using an LSTM to get contextualized hidden vectors of each word $h_i$, and use them to learn attentions by introducing another trainable vector $s$ of the same dimension of $h_i$. We then apply the attention to summarize the word embeddings into a clause embedding. Detailed equations are provided in section ~\ref{sec:lstm_attention}. 
The dashed circle in Figure~\ref{fig:overview} illustrates our LSTM-Attention based clause encoder. 

\noindent\textbf{Sentence-level Sequence Tagging.}
We observe that the discourse labels have a clear transition of logic flow (e.g. \emph{result} usually followed by \emph{implication}, and \emph{method} usually followed by \emph{hypothesis}). 
Therefore, we extend LSTM sequence tagger used by \newcite{dasigi2017experiment} to BiLSTM-CRF sequence tagger \cite{huang2015bidirectional} to label discourse types for each sentence in a paragraph. 

\noindent\textbf{Labels in BIO Scheme.}
We use the BIO scheme \cite{sang1999representing} to train all of our models (Baseline models for SciDT dataset do not use BIO2 scheme). Specifically, we convert the labels into BIO scheme where \emph{none} label represents \emph{O} and all other labels are converted into \emph{B\_label} when the previous label type is different from the current label and \emph{I\_label} when the previous label is the same as the current label. 

\subsection{Downstream Applications} 
\subsubsection{Claim Extractor}
Due to the similar problem formulation of evidence extraction task \cite{claim-extraction}, we directly employ the discourse tagging model for claim extraction. 
\subsubsection{Evidence Fragment Detector}
\label{sec:methods_evidence_fragment}
\paragraph{Problem Reduction.}
As Figure \ref{fig:evidence_fragment} shows, since each clause in evidence fragment detection task may refer to more than one subfigure codes, we cannot directly solve it as a standard classification task. Instead, we reduce it to a clause-level sequence tagging problem under a block-based assumption. We treat each paragraph as a single input. During training, we encode the subfigure code reference sequences of the clauses in each paragraph into a single BIO \cite{sang1999representing} sequence (where \emph{B} indicates the clause is the beginning of a block, \emph{I} indicates the clause is in the same block as the previous clause, and \emph{O} indicates that no subfigure code is being referred to) as demonstrated at the end of each clause in Figure \ref{fig:evidence_fragment}. For prediction, we decode the semantic subfigure code references of all clauses from the BIO sequence for each paragraph following the same block-based assumption. 

\noindent\textbf{Block-Based Assumption.}
Most subfigure code reference labels are block-based. We call contiguous clauses that share the same subfigure code reference labels as a block, which is segmented by red lines in Figure \ref{fig:evidence_fragment}. We further observe that most blocks explicitly mention all of the semantically referred subfigure codes at least once. Therefore, assuming this property is true for all blocks, we can reconstruct a sequence of semantic subfigure code references for all clauses in a paragraph. We use the explicitly mentioned subfigure codes for each block and a BIO sequence indicating where each block starts and ends for the reconstruction. Consequently, during encoding, we convert annotated semantically referenced subfigure code labels into BIO scheme. During decoding, we first localize the start and end position of each block using BIO predicted tags, then fill each block with all explicitly mentioned subfigure codes.

\noindent\textbf{Clause-level Sequence Tagger.}
The key part of our sequence tagging-based solution for evidence fragment detection is to determine where a block starts and ends. We apply a clause-level sequence tagger to tag each clause in a paragraph. Due to the small size of the evidence fragment detection dataset, we empirically observe that feature-based CRF sequence taggers outperform neural-network based sequence taggers, we thus adopt the feature-based model. In addition to the scientific discourse tags, we use explicitly mentioned subfigure codes as well as unigram, bigram and trigram words as features. For each clause, we use all features described previously from the current clause in addition to the same sets of features from the adjacent previous and next clauses.

\section{Experimental Setup}

We evaluate the performance of our scientific discourse tagger on PubMed-RCT dataset \cite{PubMed-RCT} and SciDT dataset \cite{burns2016automated, dasigi2017experiment} (Section \ref{sec:discourse_tagging_result}). We also examine the transferablity of our scientific discourse tagger to new datasets using CODA-19 dataset \cite{huang2020coda} (Section \ref{sec:coda_result}). We further study the efficiency of scientific discourse tags on claim-extraction task via transfer learning as well as evidence fragment detection task in a pipeline fashion (Section \ref{sec:downstream}).  

\subsection{Datasets}
Figure \ref{fig:count} shows the distribution of the labels in the three datasets introduced below as well as their mappings used for zero-shot predictions in Section \ref{sec:coda_result}.

\begin{figure}
  \begin{minipage}[]{0.45\textwidth}
    \begin{tabular}{cc}
  \centering
    \includegraphics[width=\textwidth]{figs/dataset_count.pdf}
   \end{tabular} 
   \end{minipage}
    \caption{Count of each label in three datasets. The lines correspond to the mappings from SciDT dataset \cite{burns2016automated,dasigi2017experiment} and PubMed 20k RCT dataset \cite{PubMed-RCT} to CODA-19 dataset \cite{huang2020coda} for zero-shot predictions (Section \ref{sec:coda_result}).} \label{fig:count}
    \vspace{-1em}
\end{figure}

\noindent\textbf{PubMed-RCT Dataset.}
We use PubMed-RCT \cite{PubMed-RCT} as the standard dataset to evaluate our scientific discourse tagger against other strong baselines. PubMed-RCT is derived from PubMed for sequential sentence classification. It has two versions -- a smaller PubMed 20k RCT, and a 10 times larger PubMed 200k RCT. Due to our limited availability of computational resources, we only consider PubMed 20k RCT in this work. PubMed 20k RCT is a large dataset that consists of 20k abstracts of randomized controlled trials (RCTs), with vocabulary of 68k across 240k sentences. Each sentence of an abstract is labeled with one of the following roles (section heads) in the abstract: \emph{background}, \emph{objective}, \emph{method}, \emph{result} or \emph{conclusion}.

\noindent\textbf{SciDT Dataset. }
Similar to PubMed-RCT \cite{PubMed-RCT}, SciDT dataset \cite{burns2016automated,dasigi2017experiment} is a clause-based dataset with more fine-grained taxonomy. We further expand SciDT dataset by applying the same clause parsing and annotation pipeline described by \newcite{dasigi2017experiment}. This dataset is derived from the Pathway Logic \cite{eker_pathway_2002} and INTACT databases \cite{orchard2013mintact}. Texts from all sections of each of those papers were pre-processed by parsing each sentence to generate a sequence of main and subordinate clauses using Stanford Parser \cite{socher2013parsing}. Domain experts were asked to label each of the clauses using the 7-label taxonomy proposed by \newcite{de2012epistemic} whose distributions are shown in Figure \ref{fig:count}. We apply sequential methods to sequences of clauses in individual paragraphs. 

Overall, SciDT dataset has a total of 634 paragraphs and 6124 clauses. We randomly split 570 paragraphs as the training and validation set and the rest as the test set. Each paragraph contains up to 30 clauses and the number of word per clause has a mean of 17.7 and a standard deviation of 12.5. The total vocabulary size is 8563, which is a small dataset for an NLP task. However, we note the difficulties of obtaining such dataset. We further perform a quality assessment of the dataset by re-annotating the test set. We obtain Cohen's kappa coefficient $\kappa=0.823$, which indicates a high quality of the dataset. 

\noindent\textbf{CODA-19 Dataset.}
CODA-19 \cite{huang2020coda} is a human-annotated dataset on a subset of the abstracts of CORD-19 \cite{wang2020cord}, which is a corpus of scholarly articles about COVID-19. \citet{wang2020cord} segmented each abstract into sentence fragments by comma (,), semicolon (;), and period (.). Each sentence fragment is labeled with one of the research aspects: \emph{background}, \emph{purpose}, \emph{method}, \emph{finding/contribution} or \emph{other}, which is similar to the label sets of PubMed-RCT \cite{PubMed-RCT}. There are 10966 abstracts in total. We use this dataset to further examine our scientific discourse tagger architecture's applicability to new datasets as well as the transferability of our trained scientific discourse tagger to new datasets.

\subsection{Baseline Models}
\noindent\textbf{PubMed-RCT Dataset.}
We compare our discourse tagger against two strong baselines on the PubMed 20k RCT dataset: (1) a hierarchical sequential labeling network (HSLN) proposed by \newcite{HSLN} and (2) the state-of-the-art model~\cite{srivastava2019hierarchical} on this dataset. HSLN~\cite{HSLN} used bio-word2vec \cite{moen2013distributional}, a word2vec embedding \cite{mikolov2013efficient} trained on corpora of Wikipedia, PubMed, and PMC, a convolutional neural network (CNN) \cite{lecun2015deep} (HSLN-CNN) or a BiLSTM (HSLN-RNN) as a sentence encoder, followed by a BiLSTM-CRF architecture \cite{huang2015bidirectional} as a sentence-level sequence tagger. 
\newcite{srivastava2019hierarchical} used a similar architecture: bio-word2vec \cite{moen2013distributional} as word embedding, BiLSTM layer with a special dilation mechanism and a capsule layer \cite{hinton2011transforming} as the sentence encoder and BiLSTM-CRF \cite{huang2015bidirectional} as the sentence-level sequence tagger.

\noindent\textbf{SciDT Dataset.}
In addition to the model of \newcite{dasigi2017experiment}
trained on our expanded SciDT dataset, we also compare with feature based CRF 
and SVM
with unigram, bigram and trigram words in the previous, current and next clauses as features. 

\noindent\textbf{CODA-19 Dataset.}
\citet{huang2020coda} composed the CODA-19 dataset and studied a few baselines for scientific discourse tagging. Their best model is a fine-tuned SciBERT~\cite{beltagy2019scibert}.

\section{Experimental Results}

\begin{table}[t]
\small
\begin{center}
    \begin{tabular}{l l l l}
    \hline
    \textbf{Model} & & \textbf{RCT} & \textbf{SciDT} \\ \hline
    \multicolumn{2}{l}{CRF} & & 0.679 \\
    \multicolumn{2}{l}{SVM} & & 0.737  \\ 
    \multicolumn{2}{l}{\newcite{dasigi2017experiment}} & & 0.791 \\ \hline
    \multicolumn{2}{l}{HSLN-CNN} & 0.922 & \\
    \multicolumn{2}{l}{HSLN-RNN} & 0.926 & \\ 
    \multicolumn{2}{l}{\newcite{srivastava2019hierarchical}} & 0.928 & \\ \hline
    \textbf{Embedding} & \textbf{Attention} &&\\ \hline
    BioGloVe & No Context  & 0.901 & 0.745 \\
    BioGloVe & RNN  & 0.909 & 0.763 \\
    BioGloVe & LSTM  & 0.913 & 0.794 \\ \hline
    BioBERT & No Context  & 0.909 & 0.794 \\
    BioBERT & RNN  & 0.915 & 0.775 \\
    BioBERT & LSTM  & 0.927 & 0.794 \\ \hline
    SciBERT & No Context  & 0.918 & 0.806 \\
    SciBERT & RNN  & 0.922 & 0.817  \\
    \textbf{SciBERT} & \textbf{LSTM}  & \textbf{0.951} & \textbf{0.841} \\ \hline
    \end{tabular}
    \vspace{1em}
    \caption{Scientific discourse tagging performance measured by test F1 score on PubMed 20k RCT and SciDT dataset.} 
    \label{tab:discourse_tagging_performance}
    \vspace{-1em}
\end{center}
\end{table}

\subsection{Supervised Learning Results} \label{sec:discourse_tagging_result}
Table \ref{tab:discourse_tagging_performance} reports the test F1 score of our scientific discourse tagger and its variations against baseline models on PubMed 20k RCT dataset and SciDT dataset. Our best scientific discourse tagger outperforms the state-of-the-art model \cite{srivastava2019hierarchical} on PubMed 20k RCT dataset by more than 2 \% absolute F1 score. Given the large size of PubMed 20k RCT, this result robustly demonstrates the strength of our model. Our model also significantly outperforms \newcite{dasigi2017experiment} with 5\% absolute F1 score (per McNemar's test, $p<0.01$). Based on these performance, we claim our scientific discourse tagger as state-of-the-art. Note that for scientific discourse tagging, the micro F1 performance is equivalent to accuracy.

\paragraph{Ablation Studies.} 
We also perform ablation studies to compare the effect of different word embeddings and attention mechanisms to the performance of our scientific discourse tagger on PubMed-RCT and SciDT dataset in Table \ref{tab:discourse_tagging_performance}. All neural network based models discussed for scientific discourse tagging tasks, including ours consist of a word embedding, a word-to-sentence encoder and a sentence-level sequence tagger. As we introduce in Section \ref{sec:methods_sciDT}, our best model has SciBERT \cite{beltagy2019scibert} as our contextualized word embedding, an LSTM-attention structure as our word-to-sentence encoder and BiLSTM-CRF \cite{huang2015bidirectional} as our sentence sequence tagger. Comparing to other baseline models, we improve the model design by adopting the state-of-the-art BERT \cite{devlin2018bert} based language model as our contextualized embedding. Instead of bidirectional LSTM as word-to-sentence encoder used by \newcite{HSLN} and \newcite{srivastava2019hierarchical}, we improve the attention structure proposed by \newcite{dasigi2017experiment}. We compare the effect of different embeddings and attention types used in scientific discourse tagger. As Table~\ref{tab:discourse_tagging_performance} indicates, our main improvement comes from SciBERT \cite{beltagy2019scibert}. In addition to BioBERT \cite{10.1093/bioinformatics/btz682} which trains BERT \cite{devlin2018bert} on biomedical domain corpus, SciBERT uses a domain specific vocabulary. BERT as a contextualized embedding also contributes partially to the performance improvement as the BioBERT embedding globally outperforms BioGloVe \cite{burns2019building}, which is a static embedding trained on biomedical domain corpus, on PubMed-RCT dataset. Another source of improvement comes from the attention structure. Our LSTM-attention outperforms the RNN-attention that \newcite{dasigi2017experiment} used.

\begin{figure}
\centering
    \includegraphics[width=0.35\textwidth]{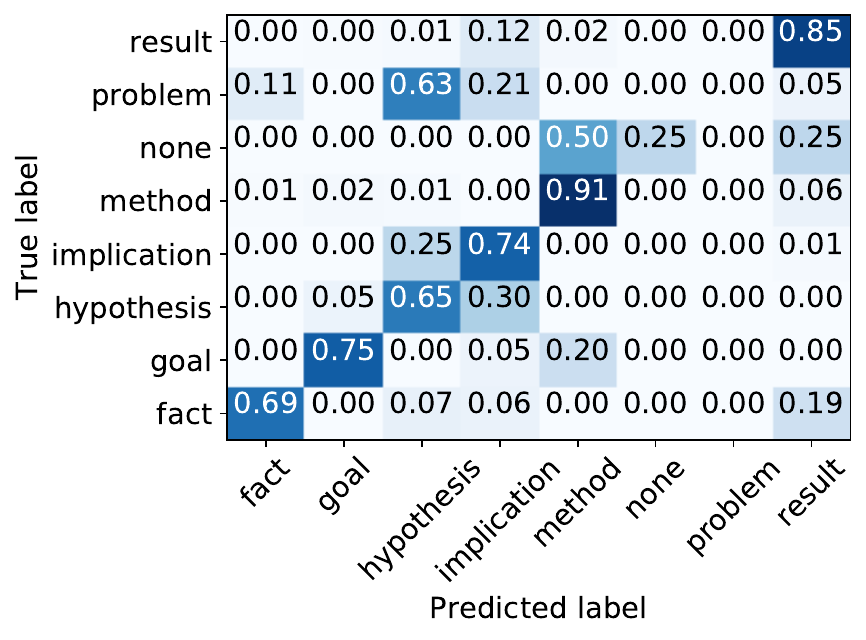}
    \includegraphics[width=0.35\textwidth]{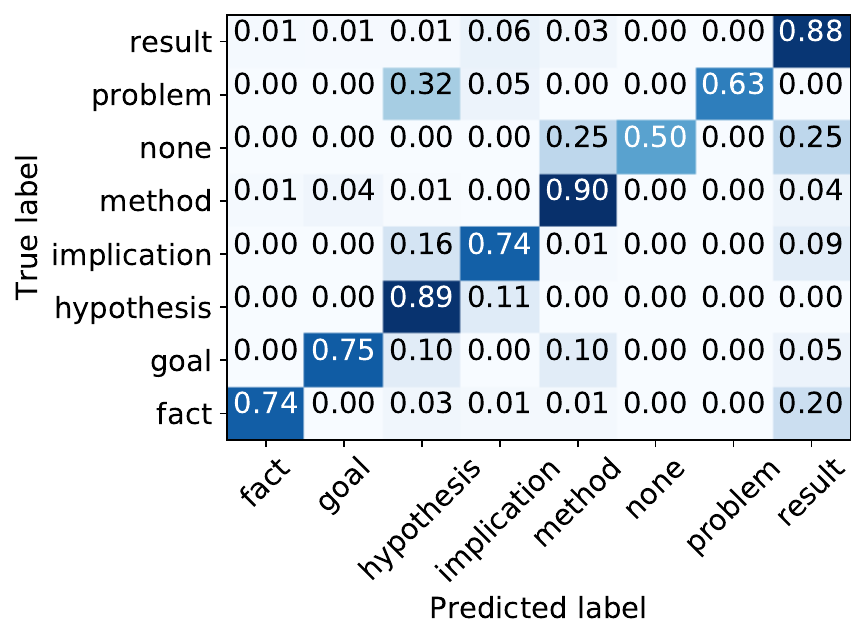}
     \vspace{-.5em}
  \caption{Confusion matrix on SciDT test data. Up: \newcite{dasigi2017experiment}. Down: Our scientific discourse tagger.} 
  \label{fig:SciDT_SciBERT_confusion}
  \vspace{-0.5em}
\end{figure}

\paragraph{Error Analysis.}
Figure 
~\ref{fig:SciDT_SciBERT_confusion} compares the confusion matrices of \newcite{dasigi2017experiment} and our best scientific discourse tagger on SciDT test set. As suggested by the overall performance, our model globally predicts the discourse tags more precisely than \newcite{dasigi2017experiment}. Specifically, \newcite{dasigi2017experiment} failed to predict \emph{problem} tag, but our model achieved 0.63 accuracy on predicting \emph{problem} tag. Figure~\ref{fig:SciDT_SciBERT_confusion} also indicates the different difficulties of predicting different discourse labels due to the imbalance of the label distributions, as Table \ref{fig:count} shows.

\subsection{Transfer Learning on CODA-19 Dataset} \label{sec:coda_result}
We further demonstrate the strong performance of our scientific discourse tagger by training it on CODA-19 dataset \cite{huang2020coda}. As Table \ref{tab:CODA_performance} shows, our model outperforms the baseline from \citet{huang2020coda} by 14.6\% absolute F1 on the test set. 

More importantly, we use these results as baselines and CODA-19 dataset as an example dataset to show the transferability of our model to new datasets. We first perform zero-shot prediction using our best \emph{trained} scientific discourse taggers on PubMed-RCT \cite{PubMed-RCT} or SciDT dataset \cite{dasigi2017experiment}. We map the labels from the original datasets to the target CODA-19 dataset by applying majority vote to the predicted labels on the training set as the lines in Figure \ref{fig:count} show. Then we perform predictions using the best trained scientific discourse taggers on CODA-19 test set and convert the predicted labels from the original label sets to the target CODA-19 label set. As a result, as Table \ref{tab:CODA_performance} shows, our zero-shot prediction results are even higher than the baseline from \citet{huang2020coda} which was directly trained on the CODA-19 dataset. This result indicates the strong transferability of our trained scientific discourse tagger as a useful tool on new datasets.

Furthermore, we separately perform a standard transfer learning by taking the scientific discourse tagger pre-trained on PubMed-RCT dataset and fine-tuning it on CODA-19 dataset. We replace the last CRF layer with a new one to match the labels of CODA-19 dataset. As a result, we achieved 0.909 test F1, which is another 2.4\% absolute F1 improvement on our model directly trained on CODA-19 dataset. This is likely due to the similar label structures between PubMed-RCT and CODA-19 dataset.

\begin{table}
\small
\begin{center}
    \begin{tabular}{l  l }
    \hline
    \textbf{Model} & \textbf{Test F1} \\ \hline
    \newcite{huang2020coda} & 0.749  \\ 
    \textbf{Ours} & \textbf{0.885} \\ \hline 
    Zero-shot Prediction from RCT & 0.760\\
    \textbf{Zero-shot Prediction from SciDT} & \textbf{0.761} \\
    \textbf{PubMed-RCT pre-train} &  \textbf{0.909} \\ \hline
    \end{tabular}
    \vspace{1em}
    \caption{Transfer Learning Performance on CODA-19 Dataset.} \label{tab:CODA_performance}
    \vspace{-1em}
\end{center}
\end{table}

\begin{table}[t]
\small
\begin{center}
    \begin{tabular}{ l l }
    \hline
    \textbf{Model} & \textbf{Test F1} \\ \hline
    \newcite{claim-extraction} & 0.790  \\ \hline
    Ours (No pre-train) & 0.791  \\
    \textbf{Ours (PubMed-RCT pre-train)} & \textbf{0.828}  \\ \hline
    \end{tabular}
     \vspace{1em}
    \caption{Claim extraction performance measured by binary F1 score, which regards $0$ as negative label.} \label{tab:claim_extraction_performance}
    \vspace{-1em}
\end{center}
\end{table}

\section{Downstream Applications} \label{sec:downstream}

\subsection{Claim Extraction}
\noindent\textbf{Dataset.}
\newcite{claim-extraction} introduced an expertly annotated dataset for extracting claim sentences from biomedical paper abstracts. They followed the definitions by \newcite{sateli2015semantic} to annotate a claim as a statement that either declares something is better, proposes something new, or describes a new finding or a causal relationship. Each sentence is tagged with a binary label indicating it is a claim or not. Each abstract may contain multiple claims as Figure \ref{fig:claim_example} shows. The dataset contains 1500 abstracts sampled from MEDLINE database.

\noindent\textbf{Baseline Model.}
\newcite{claim-extraction} constructed claim-extraction dataset and proposed a model using the sentence classification technique presented by \newcite{arora2016simple} as sentence encoding method, and the standard BiLSTM-CRF \cite{huang2015bidirectional} as the sentence-level sequence tagger. Their best model was pre-trained on PubMed 200k RCT \cite{PubMed-RCT} for transfer learning and used GloVe \cite{pennington2014glove} as their word embedding.

\noindent\textbf{Model Performance.}
Table~\ref{tab:claim_extraction_performance} compares the test binary F1 performance of \newcite{claim-extraction} with our test performance. We first train our scientific discourse tagger model directly on the claim-extraction dataset. We obtain test binary F1 score of 0.791, which is already higher than \newcite{claim-extraction}. Then as \newcite{claim-extraction} suggested, we pre-train the scientific discourse tagger on PubMed 20k RCT \cite{PubMed-RCT} and fine-tune it on the claim-extraction dataset. We replace the last CRF layer with a new one to match the binary label structure of claim-extraction dataset. As a result, we obtain test binary F1 score of 0.828, which is another 3.7\% absolute F1 improvement on our model without transfer learning. This result demonstrates the benefit of transfer learning from scientific discourse tagging task to it's downstream-tasks.


\subsection{Evidence Fragment Detection}
\noindent\textbf{Dataset.}
\newcite{burns2017extracting} introduced evidence fragment detection dataset, which shares the same format and source of clause-based paragraphs with SciDT dataset \cite{dasigi2017experiment}. As Figure \ref{fig:evidence_fragment} shows, each clause was annotated with subfigure codes that it is semantically referring to. Each clause may not refer to any subfigure code, or simultaneously refer to multiple subfigure codes. The explicit mentions of the subfigure codes were also annotated. All paragraphs are from \emph{Results} section of experimental papers, and most of the paragraphs are from a subset of SciDT dataset \cite{burns2016automated, dasigi2017experiment}. We further expand evidence fragment detection training set by annotating extra \emph{Results} section paragraphs from SciDT dataset. Overall this small dataset consists of 191 paragraphs as training data and 19 paragraphs as test data.

\noindent\textbf{Baseline Model.}
\newcite{burns2017extracting} proposed a rule-based method for evidence fragment detection task. The key steps are determining where each evidence fragment begins and ends based on the discourse tags of each clause. They treat \emph{hypothesis}, \emph{problem} and \emph{fact} as indicators of beginning of a evidence fragment, and \emph{result} and \emph{implication} as indicators of the end of a evidence fragment. They also used other features including section headings and whether the references to subfigures are entirely disjoint. Note that their document-level  rule-based tagging is across multiple paragraphs in the \emph{Results} section.

\begin{table}[t]
\small
\begin{center}
    \begin{tabular}{ c c c}
    \hline
    \textbf{Model} & \textbf{BIO F1} & \textbf{Test F1} \\ \hline
    \newcite{burns2017extracting} & N/A & 0.75  \\ \hline
    Ours (W/O Discourse Tags) & 0.750 & 0.742  \\
    \textbf{Ours (W/ Discourse Tags)} & \textbf{0.821} & \textbf{0.807}  \\ \hline
    \end{tabular}
     \vspace{1em}
    \caption{Evidence fragment detection performance measured by micro F1 score. Our block-based decoding method achieves 0.94 F1 using ground truth BIO sequences.} \label{tab:figure_span_performance}
    \vspace{-1em}
\end{center}
\end{table}

\noindent\textbf{Model Performances.}
We use a feature-based CRF 
with block-based encoding-decoding method to solve this task as a sequence tagging problem. The decoding method described in Section \ref{sec:methods_evidence_fragment} achieves 0.94 F1 score given the ground truth BIO sequences.
Table~\ref{tab:figure_span_performance} compares our feature-based CRF model performance with \newcite{burns2017extracting} 
We also compare our feature-based CRF model performances trained with or without scientific discourse tags from SciDT dataset. Our feature-based CRF model without scientific discourse tags as inputs does not outperform \newcite{burns2017extracting}. However, by adding the scientific discourse tag as a feature, we obtain 5.7\% absolute F1 improvement over \newcite{burns2017extracting}, reaching 0.807 test F1. This improvement is because of the improvement of the CRF sequence tagger. This result shows the strong benefit of scientific discourse tags as the upstream task of evidence fragment detection.

\section{Discussion}

We use the claim-extraction task and the evidence fragment detection task as two examples to demonstrate the benefit of leveraging pre-trained scientific discourse taggers and scientific discourse tags to improve the downstream-task performance via transfer learning or in a pipeline fashion. As \newcite{burns2017extracting} proposed, given the output of evidence fragment detection system, we can link subfigure codes with INTACT \cite{orchard2013mintact} records to obtain evidence fragments for each experimental figure. 

We further suggest that the evidence fragment detection task can help biocurators delineate evidence fragments as independent documents so they can be cataloged, indexed, and reused. Traditionally scientists' arguments are based on relationships between claims and evidences within the same paper and possibly a limited number of cited papers. With the help of evidence fragments, we are able to discard the convention of only linking claims to evidence from a single paper or of following citations, which are often based on linking separate claims from different papers. As a future work, we can surface the evidence fragments combined with figures and captions across multiple papers. \citet{clark2014micropublications} proposed the ``Micropublications'' semantic model, which is an abstract framework that integrates scientific argument and evidence from scientific documents. Our scientific discourse tagger, claim extractor and evidence fragment detector may serve as the actual implementation of the modules in such a framework. Ultimately, we hope to dramatically increase the amount of primary evidence used to generate individual claims and therefore improve the quality of those claims.

\section{Conclusions} 
We develop a state-of-the-art model for scientific discourse tagging and demonstrate its strong performance on PubMed-RCT dataset \cite{PubMed-RCT} and SciDT dataset \cite{burns2016automated, dasigi2017experiment} as well as its strong transferability on new datasets such as CODA-19 dataset \cite{huang2020coda}. We then demonstrate the benefit of leveraging the scientific discourse tags on downstream-tasks by providing claim-extraction task and evidence fragment detection task as two show cases. We further propose a future direction that scientific discourse tagging helps delineate evidence fragments as independent documents so they can be cataloged, indexed, and reused. As a result, we can dramatically increase the amount of primary evidence used to generate individual claims and therefore improve the quality of those claims.

\section*{Acknowledgement}
We thank the anonymous reviewers for their useful comments, and the PlusLab members for their initial feedback. This work is supported by a National Institutes of Health (NIH) R01 grant (LM012592). The views and conclusions of this paper are those of the authors and do not reflect the official policy or position of NIH.

\clearpage
\bibliography{anthology,eacl2021}
\bibliographystyle{acl_natbib}

\clearpage
\appendix

\section{Appendices}
\label{sec:appendix}

\subsection{Sentence Representations via Attention}
Each of the word representations in the input tensor $D$ is first reduced from $d$ to $d_{2}$ dimensions. Then the word representations are projected from $d_{2}$ dimension to $p$ dimension. For attention without context, $p$ dimensional reduced word representations directly perform dot product with a $p$ dimensional vector to obtain attention scores without using RNN. For attention with context, a simple RNN or LSTM with unit size $h$ is used to compute attention scores. After obtaining the summarized matrix $D_{summ}$, we use bidirectional LSTM with hidden state of size $H$ to tag the clauses.

\paragraph{LSTM-Attention} 
\label{sec:lstm_attention}
We take the input tensors $D$ of shape $c\times w\times d$ and output a matrix $A$ of shape $c\times w$ which contains the attention weights of all the words in each clause or sentence. We first project each input word into a lower dimensional space using a projection matrix $P$ of shape $d\times p$. 

{\small
\begin{equation*}
D_{l}=tanh(D \cdot P) \in  \mathbb{R}^{c\times w\times d}
\end{equation*}
}
We score $D_{l}$ with context that is summarized by an LSTM.
Specifically, we score each word in the $i^{th}$ clause in the context of other words in the same clause or sentence using an LSTM. The score for each word is a function of its $p$ dimensional representation $W_{j}$ and the previous words in the clause represented by the hidden states ($h^{i}_{j-1}$) in the LSTM cell. The equations are the following:
\begin{equation*}
\small
\begin{aligned}
    D^{i}_{l} = D_{l}[i,:,:] \in  \mathbb{R}^{w\times p} \\
    W_{j} = D^{i}_{l}[j,:] \in  \mathbb{R}^{p} \\
    h^{i}_{j} = LSTM(W_{j},\:h^{i}_{j-1}) \in  \mathbb{R}^{h} \\
    h^{i} = [h^{i}_{1}\:h^{i}_{2}\:...\:h^{i}_{w}] \in  \mathbb{R}^{w \times h} \\
    a^{i} = softmax(h^{i} \cdot s) \in  \mathbb{R}^{w} \\
    A = [a^{1}\:a^{2}\:...\:a^{i}\:...\:a^{c}] \in  \mathbb{R}^{c \times w} \\
\end{aligned}
\end{equation*}
where $LSTM$ is an LSTM cell with the unit size of $h$. $s$ is a vector of length $h$.

Finally like \newcite{dasigi2017experiment}, a $c \times d$ shaped weighted sum $D_{summ}$ of the input tensor $D$ is computed, with the weights computed by the attention mechanism, then it is fed to a clause/sentence-level sequence tagger to tag discourse labels.

{\small
\begin{equation*}
    D_{summ}[i,:] = A[i,:] \cdot D[i,:,:] \in  \mathbb{R}^{d}
\end{equation*}
}

\subsection{Implementation and Training Details}
The scientific discourse tagging model is implemented using Keras \cite{chollet2015keras} with Tensorflow \cite{tensorflow2015-whitepaper} backend. We use early stopping mechanism with toleration of 2 epochs. We schedule the training by training the model with a learning rate of $lr$ for 20 epochs. We use Adam \cite{kingma2014adam} as our optimizer. The optimal hyper-parameters and the attempted range if applicable are listed in Table~\ref{tab:hyper}.

\begin{table}[t]
\begin{center}
    \begin{tabular}{  l l }
    \hline
    \textbf{Hyper-Parameter} & \textbf{Used} \\ \hline
    $d_{BERT}$ & 768 \\ \hline
    $c$ & 40  \\ \hline
    $w$ & 60  \\ \hline
    $d$ & 768 \\ \hline
    $d_{2}$ & 300 \\ \hline
    $p$ & 200 \\ \hline
    $h$ & 75 \\ \hline
    $H$ & 350 \\ \hline
    $lr$ & $10^{-3}$ \\ \hline
    Validation Set Ratio & 0.1 \\ \hline
    Embedding dropout & 0.4 \\ \hline
    Dense dropout & 0.4 \\ \hline
    Attention dropout & 0.6 \\ \hline
    LSTM dropout & 0.5 \\ \hline
    Batch size & 10 \\ \hline
    \end{tabular}
    \vspace{1em}
    \caption{Optimal hyper-parameters of scientific discourse tagger model} \label{tab:hyper}
     \vspace{-2em}
\end{center}
\end{table}


\end{document}